\documentclass{article}

\usepackage{arxiv}

\usepackage[utf8]{inputenc} 
\usepackage[T1]{fontenc}    
\usepackage{hyperref}       
\usepackage{url}            
\usepackage{booktabs}       
\usepackage{amsfonts}       
\usepackage{nicefrac}       
\usepackage{microtype}      
\usepackage{lipsum}		
\usepackage{graphicx}
\usepackage{natbib}
\usepackage{doi}
\usepackage{graphicx}
\usepackage{amsmath}
\usepackage{amssymb}
\usepackage{booktabs}
\usepackage{float}
\usepackage{makecell}
\usepackage{multirow}

\title{A Non-monotonic Smooth Activation Function}

\author{ {Koushik Biswas} \\
	\And
	{Meghana Karri} \\
       \And 
       {Ulaş Bağcı}
}

\date{}

\renewcommand{\shorttitle}{New Activation function}

\hypersetup{
pdftitle={Sqish},
pdfsubject={},
pdfauthor={Koushik Biswas},
pdfkeywords={Deep Learning, Activation Function},
}

\begin{document}
\maketitle

\begin{abstract}
Activation functions are crucial in deep learning models since they introduce non-linearity into the networks, allowing them to learn from errors and make adjustments, which is essential for learning complex patterns. The essential purpose of activation functions is to transform unprocessed input signals into significant output activations, promoting information transmission throughout the neural network. In this study, we propose a new activation function called Sqish, which is a non-monotonic and smooth function and an alternative to existing ones. We showed its superiority in classification, object detection, segmentation tasks, and adversarial robustness experiments. We got an 8.21\% improvement over ReLU on the CIFAR100 dataset with the ShuffleNet V2 model in the FGSM adversarial attack. We also got a 5.87\% improvement over ReLU on image classification on the CIFAR100 dataset with the ShuffleNet V2 model.
\end{abstract}

\section{Introduction}
\label{sec:intro}
In recent years, artificial neural networks (ANN) have emerged as the dominant paradigm in a variety of machine learning domains, and ANN can solve challenging problems, including audio and image recognition and natural language processing. The ability of ANNs to learn complex patterns and representations from unstructured input using a network of linked nodes called neurons lies at the heart of their capabilities. However, the choice and efficacy of the activation functions utilized within these networks substantially impact how effectively they learn.

Activation functions play a critical role in neural networks for introducing non-linearity into the network. Their primary role is to modulate the output of a neuron, which affects the network's capacity to replicate extremely complicated and non-linear interactions present in real-world data. Despite their apparent simplicity, activation function selection and design are critical in establishing a neural network's overall performance, convergence, and generalizability.

The notion of binary activation was initially established in McCulloch and Pitts' seminal work on neural networks in 1943, ushering in the history of activation functions. Since then, a variety of activation functions have been proposed, each with unique characteristics and benefits. Traditional activation functions such as the step, sigmoid, and hyperbolic tangent (tanh) paved the way for more recent activation functions such as the rectified linear unit (ReLU) \cite{relu} and its variants, which have gained popularity due to their ability to handle the vanishing gradient problem effectively and efficiently compare to tanh and sigmoid.

Despite the fact that there are several activation functions accessible, selecting the optimal one remains a challenge. Specific activation functions may be more suited for particular activities and network topologies. As neural networks grow in size and complexity, understanding the underlying properties and performance impact of diverse activation functions becomes increasingly crucial.

The purpose of this work is to offer a novel activation function. We will also look at how the proposed activation functions influence neural network performance in image classification, object detection, 3D medical imaging, and adversarial attack problems.

\section{Related Work}
\label{sec:relatedwork}
Activation functions play a very crucial role in ANN. It introduces the non-linear transformations that help the networks to capture intricate patterns in data. Numerous activation functions have been put forth over time, each with its own set of benefits and restrictions. 

The sigmoid activation function, introduced early in the field, was one of the first non-linearities used in neural networks. Despite its historical significance, the sigmoid function suffers from the vanishing gradient problem, which hampers training deep networks. As a response to this issue, the hyperbolic tangent (tanh) activation function was introduced, aiming to alleviate the vanishing gradient problem to some extent. However, both sigmoid and tanh functions tend to saturate for large input values, leading to slower convergence during training. 

The Rectified Linear Unit (ReLU) \cite{relu} activation function was proposed to overcome these restrictions. ReLU is computationally efficient and speeds up convergence by only activating for positive input values. However, ReLU has a major drawback, called the dying ReLU problem, where neurons can become inactive during training.

To overcome the drawbacks of ReLU, several variants of ReLU have been proposed \cite{lrelu, prelu, relu6, rlrelu}. The Leaky ReLU activation function \cite{lrelu} adds a small value for negative input values, preventing neurons from becoming entirely inactive. The Parametric ReLU (PReLU) \cite{prelu} extends this idea by making the negative slope a parameter learnable. The Exponential Linear Unit (ELU) \cite{elu} mitigates the dying ReLU problem by introducing a smooth output for negative input values and exhibiting faster convergence.

Recent breakthroughs in neural network architectures have also facilitated the development of novel smooth activation functions. In contrast to conventional ReLU-based functions, the Swish \cite{swish} activation function, found by neural architecture search \& developed by the Google Brain team, is a smooth activation function and has the potential to offer improved training dynamics. Swish blends smooth, non-linear, non-monotonic activation functions, and it can be shown that it is a smooth approximation of the ReLU function. The smoothness enhances gradients and lowers the chance of dead neurons, which helps to promote quicker and more stable convergence. GELU \cite{gelu} is another popular activation function which is recently been used in BERT\cite{bert}, and GPT \cite{gpt2, gpt3} based architectures. The vanishing gradient issue is also addressed by the GELU activation function, which encourages the training of deeper networks and makes it possible to simulate more complex data patterns. Its ability to do computations efficiently is another factor that makes it appealing for use in practical applications. The Pade Activation Unit (PAU) \cite{pau} considers approximation of known activation function by rational polynomial functions. It improves network performance in image classification problems over ReLU and Swish.

\section{Motivation}
\label{sec:motivation}
The choice of an appropriate activation function is an important problem in neural network design since it directly influences the network's expressive capability, training dynamics, and convergence features. The inspiration for this study originates from the necessity to investigate and assess the several widely used activation functions accessible in order to choose the best one for some specific deep learning tasks.

While ReLU and its variants have gained significant popularity, they have some drawbacks. The properties of an activation function can have a considerable impact on the network's capacity to simulate complicated connections in data, particularly in cases involving highly non-linear distributions. This work intends to add to the existing body of knowledge by extensively researching a novel activation function that solves the limits of current alternatives while improving deep neural network performance. We summarise the paper as follows:
\begin{enumerate}
    \item We have proposed a novel activation function that is non-monotonic and smooth. The proposed function can approximate the ReLU, Leaky ReLU function.
    \item We run extensive experiments to show the efficacy of the proposed activation function.
\end{enumerate}

\section{Proposed Method}
\label{sec:method}
We propose a novel activation function using a smooth approximation of the maximum function. The proposed function can approximate Maxout \cite{maxout}, ReLU \cite{relu}, Leaky ReLU \cite{lrelu}, or its variants. 
\subsection{Smooth Approximation}
The maximum function is defined as follows:
\begin{align}\label{eq1}
    max(x_1,x_2) &=\begin{cases} x_1 &\text{ if }x_1\geq x_2\\ x_2&\text{otherwise}\end{cases} 
\end{align}
We can rewrite the equation~(\ref{eq1}) as follows:
\begin{align}\label{eq2}
    max(x_1,x_2) &= x_1 + max(0, x_2 - x_1)
\end{align}
\begin{figure*}[!t]
\begin{minipage}[t]{.32\linewidth}
        \centering
         \includegraphics[width=\linewidth]{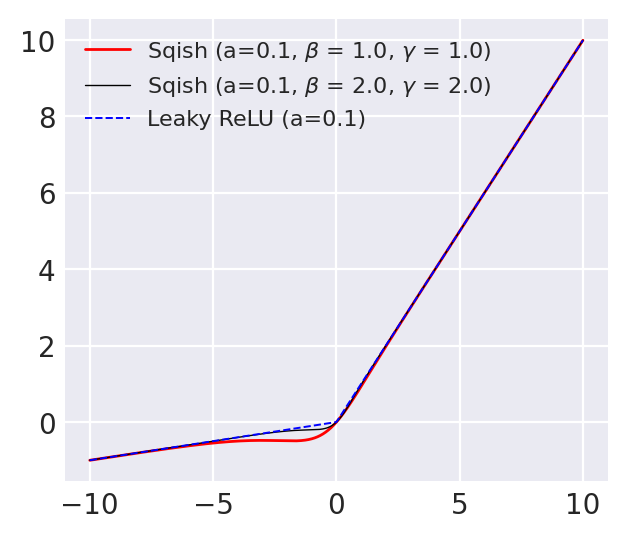}
        
        \caption{Approximation of Leaky ReLU ($a = 0.1$) using Sqish ($a = 0.1$) for different values of $\beta$ and $\gamma$. As $\gamma \rightarrow \infty$, Sqish smoothly approximate Leaky ReLU.}
        \label{sqish}
         \end{minipage}
         \hfill
   \begin{minipage}[t]{.32\linewidth}
        \centering
    
         \includegraphics[width=\linewidth]{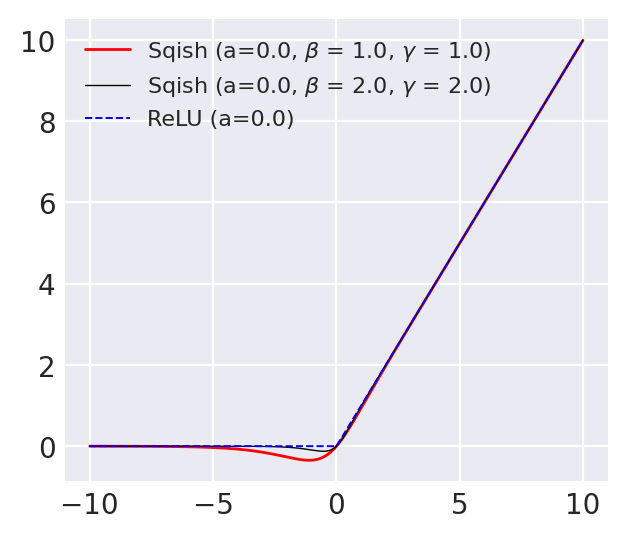}
        
        \caption{Approximation of ReLU using Sqish for different values of $\beta$ and $\gamma$. As $\gamma \rightarrow \infty$, Sqish smoothly approximates ReLU.}
        \label{sqish1}
   \end{minipage}
    \hfill
   \begin{minipage}[t]{.31\linewidth}
        \centering
    
         \includegraphics[width=\linewidth]{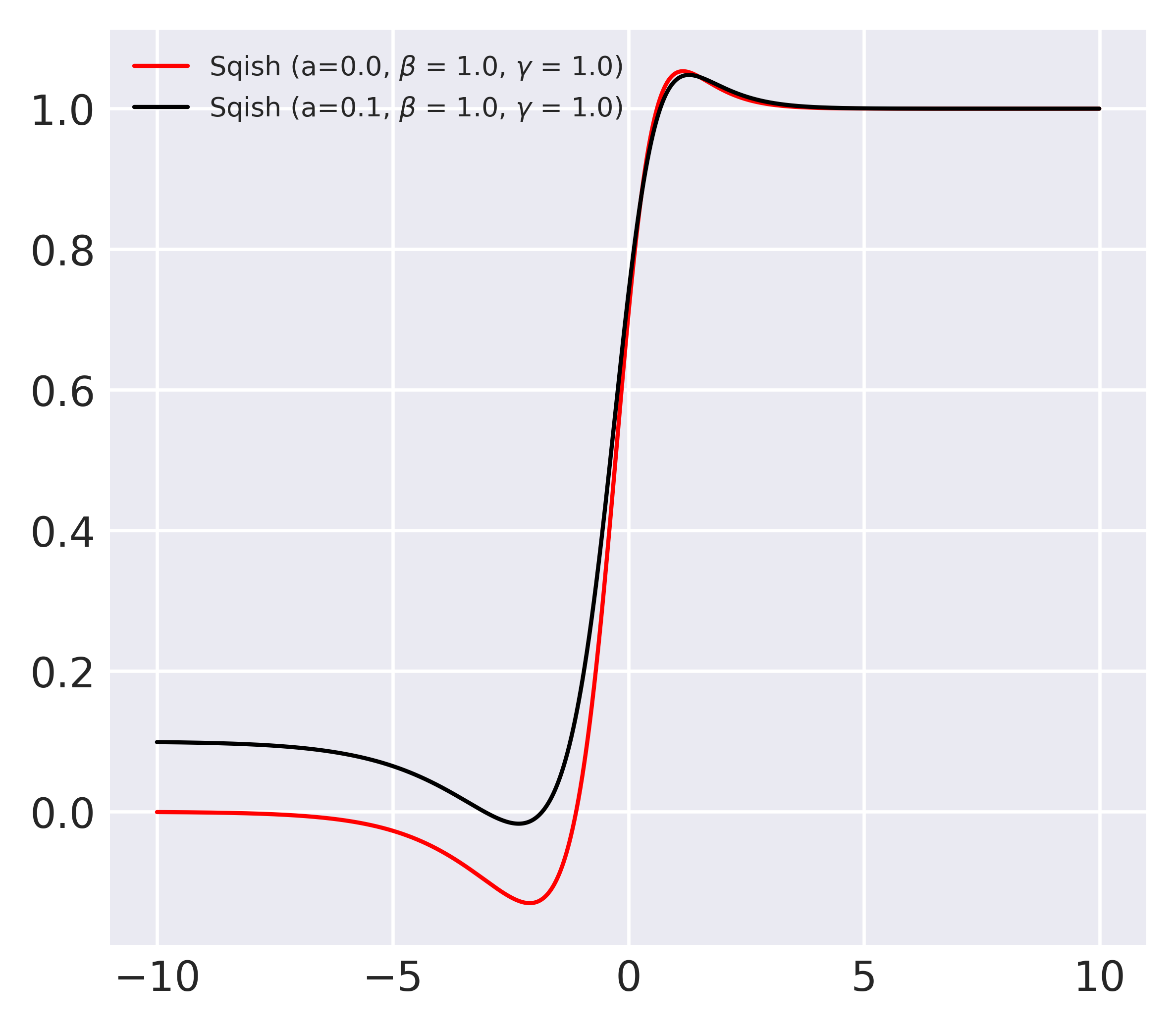}
       
        \caption{First order derivatives of Sqish with respect to 'x' for different values of $a$, $\beta$ and $\gamma$.}
        \label{der}
   \end{minipage}
\end{figure*}


An idea behind equation~(\ref{eq2}) is given in the supplementary document. It is a known fact that the maximum function is not differentiable in the real line. Now, note that the second term in equation~(\ref{eq2}) is not differentiable. We can get a smooth function from equation~(\ref{eq2}) by using an approximation of the maximum function by a smooth function. We have proposed a new approximation of the $max(0,x)$ function, which is defined as follows:
\begin{align}\label{eq3}
    F(x;\gamma) =  \dfrac{x}{\sqrt{1+  e^{-2\gamma x}}}, \ \ \ \gamma > 0
\end{align}
Note that, if we consider $\gamma \rightarrow \infty$, We have $f(x;\gamma)$ $\rightarrow$ $max(0,x)$. The function in equation~(\ref{eq3}) is a smooth function in the real line. Replacing the second term in equation~(\ref{eq2}) by the approximation proposed in equation~(\ref{eq3}), we have a pointwise approximation formula by a smooth function for the maximum function as follows:
\begin{align}\label{eq4}
    f(x_1,x_2;\gamma) &= x_1 + \dfrac{(x_2-x_1)}{\sqrt{1+  e^{-2\gamma (x_2-x_1)}}}, \ \ \ \gamma > 0
\end{align}
Note that if we consider $\gamma \rightarrow \infty$, we have $f(x_1,x_2;\gamma) \rightarrow max(x_1,x_2)$. We can get known activation functions for specific values of $x_1$ and $x_2$. Equation~(\ref{eq4}) can approximate the Maxout family. In particular, consider $x_1 = ax$ and $x_2 = bx$, we have, 
\begin{align}\label{eq5}
    f(ax,bx;\gamma) &= x_1 + \dfrac{(bx-ax)}{\sqrt{1+  e^{-2\gamma (bx-ax)}}}, \ \ \ \gamma > 0
\end{align}
This is a very simple case from the Maxout family, but we can derive a more complex formula by considering different values of $x_1$ and $x_2$. Now, considering $x_1 = 0$ and $x_2=x$, we have an approximation of the ReLU activation function by a smooth function.
\begin{align}\label{eq6}
    f(0,x;\gamma) &= \dfrac{x}{\sqrt{1+  e^{-2\gamma x}}}, \ \ \ \gamma > 0
\end{align}
Similarly, we can derive an approximation of the Leaky ReLU activation (or parametric ReLU, depending on if `a' is a hyperparameter or a trainable parameter) function by considering $x_1 = ax$ and $x_2=x$.
\begin{align}\label{eq7}
    f(ax, x;\gamma) = ax +  \dfrac{(1-a)x}{\sqrt{1+ e^{-2\gamma (1-a) x}}} , \ \ \ \gamma > 0
\end{align}
We have introduced another parameter $\beta$ in equation~(\ref{eq7}) as follows:
\begin{align}\label{eq8}
    f(ax,x;\beta, \gamma) = ax +  \dfrac{(1-a)x}{\sqrt{1+ \beta e^{-2\gamma(1-a) x}}}, \ \beta > 0, \gamma > 0
\end{align}
$\beta$ will control the function smoothness in both negative and positive axes in the real line. For the rest of the paper, we will use equation~(\ref{eq8}) as our proposed activation function, which we call \textbf{Sqish}. Figure~\ref{sqish}, ~\ref{sqish1}, and ~\ref{der} represent the plots for Sqish and its derivative for different values of $a, \beta,$ and $\gamma$.

The derivative of the proposed function in equation~(\ref{eq8}) with respect to the variable $x$ is 
\begin{align}\label{eq9}
    \frac{\partial f(x;a, \beta, \gamma)}{\partial x} = a + \dfrac{1-a}{\sqrt{{\beta}\mathrm{e}^{-2\left(1-a\right){\gamma}x}+1}}\\ \notag +\dfrac{\left(1-a\right)^2{\beta}{\gamma}x\mathrm{e}^{-2\left(1-a\right){\gamma}x}}{\left({\beta}\mathrm{e}^{-2\left(1-a\right){\gamma}x}+1\right)^\frac{3}{2}}
\end{align}
\subsection{Learning Activation parameter}
The proposed activation function in equation~(\ref{eq8}) has parameters that can be used as hyperparameters (as a fixed activation function) or trainable parameters (as a trainable activation function). Trainable parameters in a trainable activation function are updated using the backpropagation algorithm \cite{backp}. Gradients with respect to the parameters $a, \beta, \text{and} \ \gamma$ can be computed as follows:
\begin{align}\label{eq10}
  \frac{\partial f(x;a, \beta, \gamma)}{\partial a} = x -\dfrac{x}{\sqrt{{\beta}\mathrm{e}^{-2x{\gamma}\cdot\left(1-a\right)}+1}}\\ \notag -\dfrac{{\beta}x^2{\gamma}\cdot\left(1-a\right)\mathrm{e}^{-2x{\gamma}\cdot\left(1-a\right)}}{\left({\beta}\mathrm{e}^{-2x{\gamma}\cdot\left(1-a\right)}+1\right)^\frac{3}{2}}
\end{align}
\begin{align}\label{eq11}
  \frac{\partial f(x;a, \beta, \gamma)}{\partial \beta} = \dfrac{\left(a-1\right)x\mathrm{e}^{-2\left(1-a\right)x{\gamma}}}{2\left(\mathrm{e}^{-2\left(1-a\right)x{\gamma}}{\beta}+1\right)^\frac{3}{2}} 
\end{align}
\begin{align}\label{eq12}
  \frac{\partial f(x;a, \beta, \gamma)}{\partial \gamma} = \dfrac{\left(a-1\right)^2{\beta}x^2\mathrm{e}^{-2\left(1-a\right)x{\gamma}}}{\left({\beta}\mathrm{e}^{-2\left(1-a\right)x{\gamma}}+1\right)^\frac{3}{2}}
\end{align}
Note that $a$, $\beta$, and $\gamma$ can be used as hyperparameters or trainable parameters.

Now, note that the class of neural networks with Sqish activation function is dense in $C(K)$, where $K$ is a compact subset of $\mathbb{R}^n$ and $C(K)$ is the space of all continuous functions over $K$.\\
The proof follows from the following theorem.

\textbf{Theorem 1.1 from Kidger and Lyons, 2020 \cite{universal}}:- Let $\rho: \mathbb{R}\rightarrow \mathbb{R}$ be any continuous function. Let $N_n^{\rho}$ represent the class of neural networks with activation function $\rho$, with $n$ neurons in the input layer, one neuron in the output layer, and one hidden layer with an arbitrary number of neurons. Let $K \subseteq \mathbb{R}^n$ be compact. Then $N_n^{\rho}$ is dense in $C(K)$ if and only if $\rho$ is non-polynomial.

\section{Experiments:}
\label{sec:exp}
The following subsections present a detailed experimental evaluation of different deep-learning problems with standard datasets. To compare performance with Sqish, we consider eight widely used activation functions- ReLU, Leaky ReLU, PReLU, ELU, Swish, GELU, Mish, and PAU as baseline activation functions. All the experiments are conducted on NVIDIA RTX 3090 and NVIDIA Tesla A100 GPUs. Swish, PReLU, PAU, and Sqish (the proposed function) are considered trainable activation functions, and the trainable parameters are updated via the backpropagation algorithm \cite{backp}.
\subsection{Image Classification}
We report detailed experimental results on the image classification problem with MNIST \cite{mnist}, Fashion MNIST\cite{fashion}, SVHN \cite{SVHN}, CIFAR10 \cite{cifar10}, CIFAR100 \cite{cifar10}, and Tiny Imagenet \cite{tiny} datasets. Details are given in the following subsections.
\subsubsection{MNIST, Fashion MNIST, and SVHN}
In this section, we have reported results with MNIST \cite{mnist}, Fashion MNIST\cite{fashion}, and SVHN \cite{SVHN} datasets. There are a total of 60k training images and 10k test images in the $28 \times 28$ greyscale format in the MNIST and Fashion MNIST databases in 10 distinct classes. SVHN database includes 73257 training images and 26032 test images. All the images are in $32 \times 32$ RGB format with ten distinct classes. Basic data augmentation methods like zoom and rotation are applied to the SVHN dataset. We take into account a batch size of 128, 0.01 initial learning rate, and cosine annealing learning rate scheduler \cite{cosan} to decay the learning rate. We train all networks up to 100 epochs using stochastic gradient descent \cite{sgd1, sgd2} optimizer with 0.9 momentum \& $5e^{-4}$ weight decay. For the mean of 5 distinct runs, we report our results using the AlexNet architecture \cite{alexnet} in Table~\ref{app7}.
\begin{table*}[!htbp]
\centering
\begin{tabular}{ |c|c|c|c|c| }
 \hline
 Activation Function &  \makecell{MNIST} & \makecell{Fashion MNIST} & \makecell{SVHN}  \\
 
 \hline
 ReLU  &  $99.44 \pm 0.08$ & $92.68 \pm 0.21$ & $95.07\pm 0.14$\\ 
 \hline
 
Leaky ReLU & $99.44 \pm 0.09$ & $92.72 \pm 0.17$ & $95.12 \pm 0.16$\\

 \hline
 PReLU & 99.47 $\pm$ 0.08 & 92.71 $\pm$ 0.18 & 95.04 $\pm$ 0.14\\
 \hline
 ELU  & $99.49 \pm 0.06$ & $92.88 \pm 0.13$ & $95.11 \pm 0.15$\\
\hline
GELU & 99.56$ \pm$ 0.06 & 93.01 $\pm$ 0.14 & 95.12 $\pm$ 0.12\\
\hline
  Swish  & $99.55\pm 0.05$ & $92.88\pm 0.16$ & $95.28\pm 0.15$\\
 \hline
 PAU & $99.53 \pm 0.11$ & 93.10 $\pm$ 0.16 & 95.26 $\pm$ 0.17\\
\hline
Mish & $99.62 \pm 0.06$ & 93.16 $\pm$ 0.14 & 95.37 $\pm$ 0.11\\

\hline
  Sqish & \textbf{99.67} $\pm$ 0.06 & \textbf{93.35} $\pm$ 0.12 & \textbf{95.60} $\pm$ 0.11\\
 \hline
\end{tabular}
\caption{Comparison between Sqish and other baseline activations on MNIST, Fashion MNIST, and SVHN datasets for image classification problem on AlexNet architecture. We report Top-1 test accuracy (in \%) for the mean of 5 different runs. mean$\pm$std
is reported in the Table.} 
\label{app7}
\end{table*}
\subsubsection{CIFAR}
\begin{figure*}[!t]
   \begin{minipage}[t]{.47\linewidth}
        \centering
    
        \includegraphics[width=\linewidth]{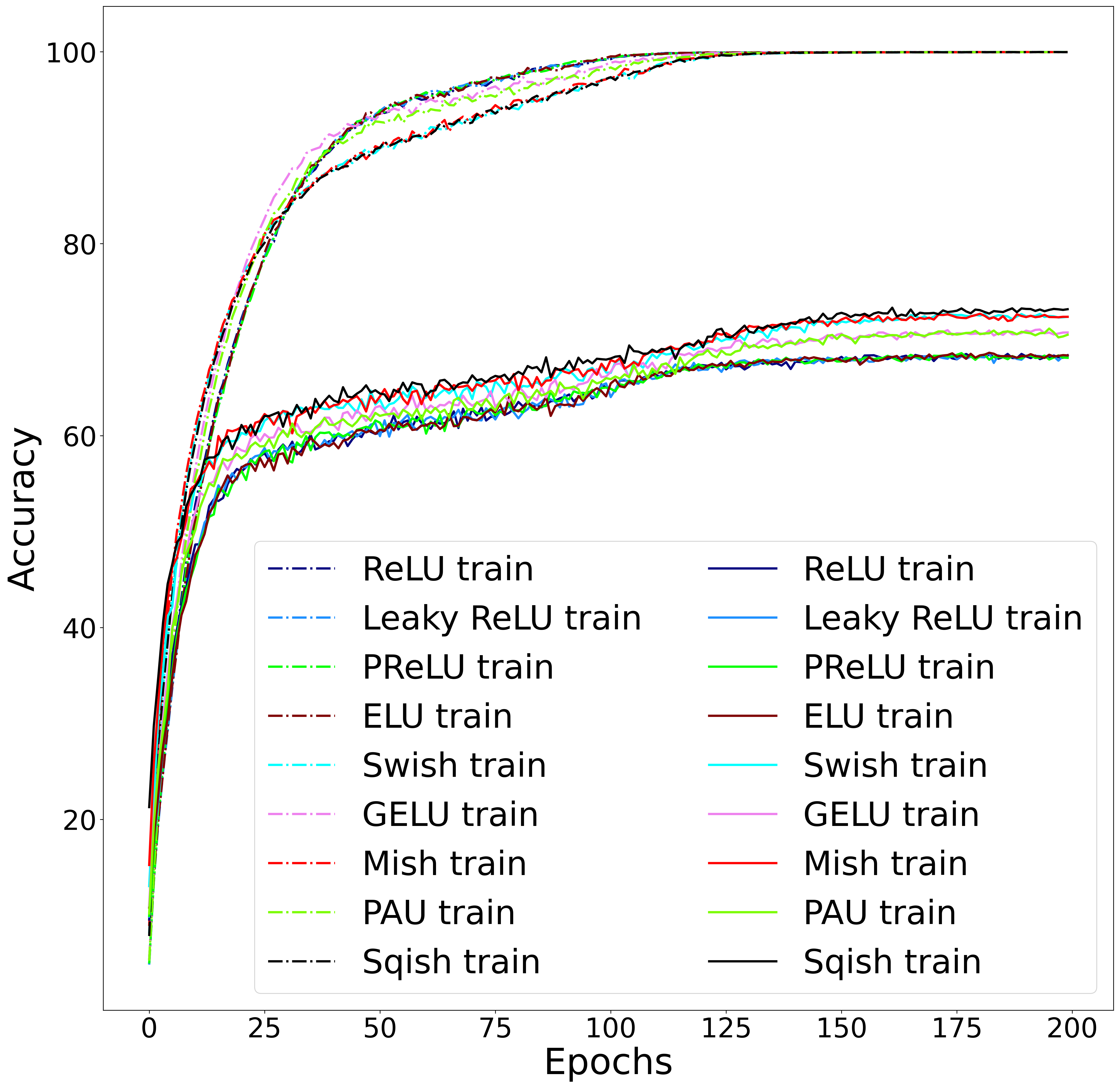}
        
        \caption{ Top-1 train and test accuracy curves for Sqish and other baseline activation functions on CIFAR100 dataset with ShuffleNet V2.}
        \label{acc2}
    \end{minipage}
    \hfill
    \begin{minipage}[t]{.46\linewidth}
        \centering
        
       \includegraphics[width=\linewidth]{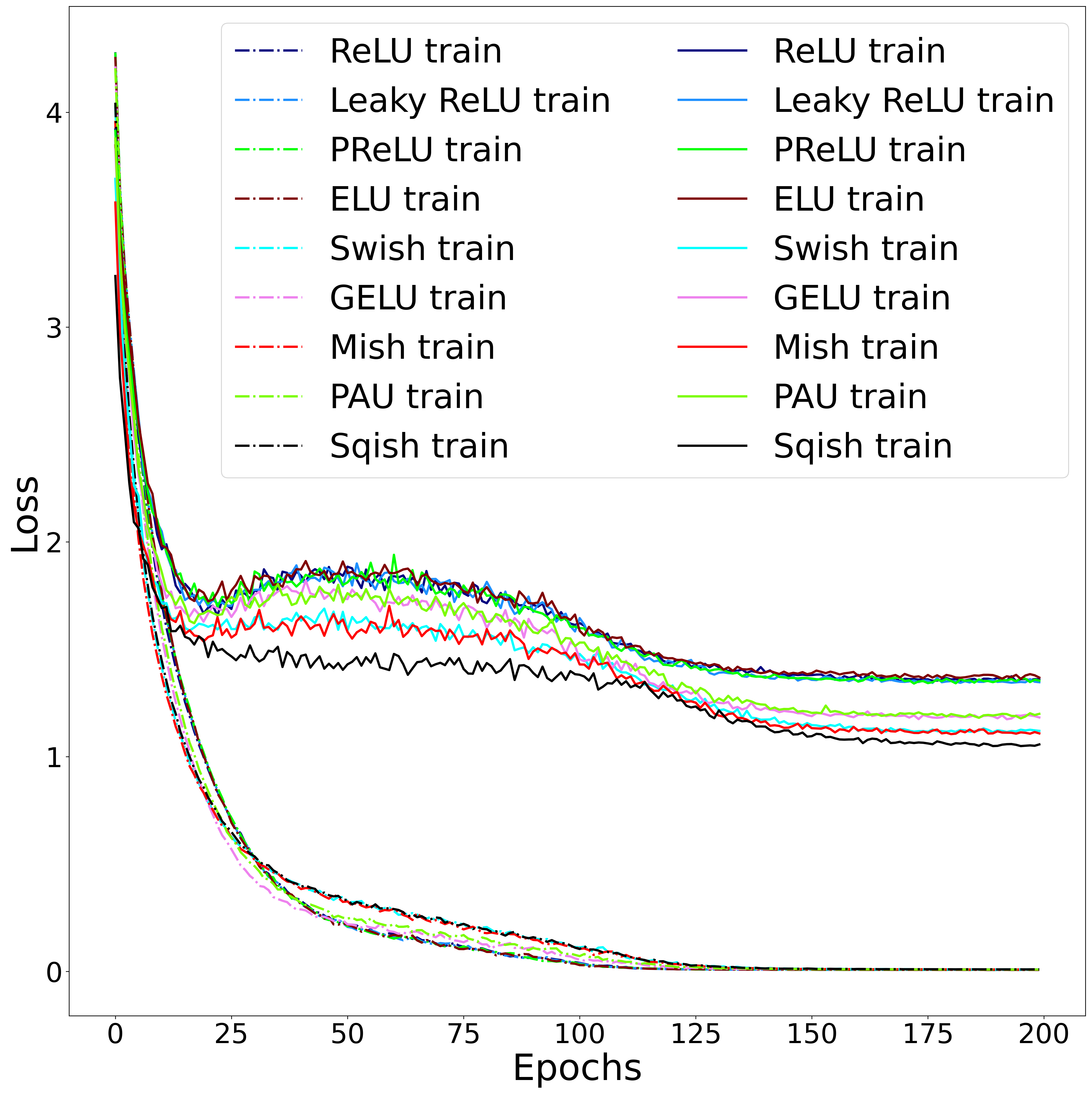}
      
        \caption{Top-1 train and test loss curves for Sqish and other baseline activation functions on CIFAR100 dataset with ShuffleNet V2.}
        \label{loss2}
    \end{minipage}  
\end{figure*}

In this section, we present results from the standard image classification datasets CIFAR10 \cite{cifar10} and CIFAR100 \cite{cifar10}.
Both datasets contain 50k training and 10k testing photos. CIFAR10 has ten classes, while CIFAR100 offers 100 classes. In these two datasets, we consider a batch size of 128, 0.01 initial learning rate, and decay the learning rate with cosine annealing learning rate scheduler \cite{cosan}. We use stochastic gradient descent \cite{sgd1, sgd2} optimizer with 0.9 momentum \& $5e^{-4}$ weight decay and trained all networks up to 200 epochs. We use data augmentation methods, including horizontal flipping and rotation. Top-1 accuracy is presented in Table~\ref{tab2} and Table~\ref{tab3} for the CIFAR10  and CIFAR100 datasets, respectively, for a mean of 10 different runs. We report results with AlexNet (An) \cite{alexnet}, LeNet (LN) \cite{lenet}, VGG-16 \cite{vgg}, WideResNet 28-10 (WRN 28-10) \cite{wrn}, ResNet (RN) \cite{resnet}, PreActResNet (PARN) \cite{preactresnet}, DenseNet-121 (DN 121) \cite{densenet}, Inception V3 (IN V3) \cite{incep}, MobileNet V2 (MN V2) \cite{mobile}, ResNext (RNxt) \cite{resnext}, Xception (XT) \cite{xception}, and EfficientNet B0 (EN B0) \cite{efficientnet}. The learning curves on the CIFAR100 dataset with the ShuffleNet V2 (2.0x) model for the baseline and Sqish are given in Figures~\ref{loss2} and ~\ref{acc2}.
\begin{table*}[!htbp]
\begin{center}
\scriptsize
\begin{tabular}{ |c|c|c|c|c|c|c|c|c|c|c|c|c|c|c| }
 \hline
\makecell{Activation\\ Function} & AN & LN & \makecell{VGG\\16} & \makecell{WRN\\ 28-10} & \makecell{RN\\50} & \makecell{PARN\\34} & \makecell{DN\\121}  & \makecell{IN\\V3} & \makecell{MN\\V2} & \makecell{SF-V2\\ 2.0x} & RNxt & XT & \makecell{EN\\B0} & \makecell{RN\\18} \\

 \hline 
 ReLU & \makecell{54.98\\$\pm$0.26} & \makecell{45.35\\$\pm$0.27} & \makecell{71.62\\$\pm$0.25} & \makecell{76.39\\$\pm$0.23} &   \makecell{74.28\\$\pm$0.22} &  \makecell{73.23\\$\pm$0.20} & \makecell{75.60\\$\pm$0.25} & \makecell{74.33\\$\pm$0.24} & \makecell{74.00\\$\pm$0.22} & \makecell{67.49\\$\pm$0.26} & \makecell{74.40\\$\pm$0.22} & \makecell{71.18\\$\pm$0.24} & \makecell{76.63\\$\pm$0.23} & \makecell{73.12\\$\pm$0.22}\\ 
 \hline
 Leaky ReLU & \makecell{55.20\\$\pm$0.26} & \makecell{45.60\\$\pm$0.25} & \makecell{71.76\\$\pm$0.26} & \makecell{76.56\\$\pm$0.24} &   \makecell{74.20\\$\pm$0.25} &  \makecell{73.35\\$\pm$0.24} & \makecell{75.79\\$\pm$0.25} & \makecell{74.32\\$\pm$0.25} & \makecell{74.20\\$\pm$0.22} & \makecell{67.61\\$\pm$0.25} & \makecell{74.50\\$\pm$0.25} & \makecell{71.10\\$\pm$0.25} & \makecell{76.89\\$\pm$0.26} & \makecell{73.05\\$\pm$0.24}\\
 \hline
  
 PReLU & \makecell{55.62\\$\pm$0.26} & \makecell{45.50\\$\pm$0.26} & \makecell{71.85\\$\pm$0.31} & \makecell{76.77\\$\pm$0.25} &   \makecell{74.40\\$\pm$0.27} &  \makecell{73.20\\$\pm$0.28} & \makecell{76.10\\$\pm$0.28} & \makecell{74.40\\$\pm$0.28} & \makecell{74.42\\$\pm$0.33} & \makecell{68.45\\$\pm$0.27} & \makecell{74.55\\$\pm$0.25} & \makecell{71.25\\$\pm$0.26} & \makecell{76.70\\$\pm$0.28} & \makecell{73.19\\$\pm$0.24}\\
 \hline
 ELU & \makecell{55.81\\$\pm$0.27} & \makecell{46.10\\$\pm$0.28} & \makecell{71.83\\$\pm$0.26} & \makecell{76.50\\$\pm$0.29} &   \makecell{74.40\\$\pm$0.23} &  \makecell{73.69\\$\pm$0.26} & \makecell{75.99\\$\pm$0.28} & \makecell{74.62\\$\pm$0.27} & \makecell{74.43\\$\pm$0.24} & \makecell{67.89\\$\pm$0.28} & \makecell{74.70\\$\pm$0.23} & \makecell{71.59\\$\pm$0.26} & \makecell{76.73\\$\pm$0.28} & \makecell{73.33\\$\pm$0.26}\\
 \hline
 Swish & \makecell{57.72\\$\pm$0.24} & \makecell{47.42\\$\pm$0.25} & \makecell{72.15\\$\pm$0.24} & \makecell{77.11\\$\pm$0.25} &   \makecell{75.19\\$\pm$0.23} &  \makecell{73.90\\$\pm$0.25} & \makecell{76.40\\$\pm$0.28} & \makecell{75.45\\$\pm$0.28} & \makecell{75.15\\$\pm$0.27} & \makecell{71.45\\$\pm$0.28} & \makecell{75.20\\$\pm$0.25} & \makecell{72.30\\$\pm$0.22} & \makecell{77.22\\$\pm$0.22} & \makecell{73.68\\$\pm$0.24}\\
 \hline
 Mish & \makecell{58.37\\$\pm$0.22} & \makecell{\textbf{47.52}\\$\pm$0.26} & \makecell{72.50\\$\pm$0.24} & \makecell{77.50\\$\pm$0.27} &   \makecell{76.43\\$\pm$0.24} &  \makecell{75.11\\$\pm$0.24} & \makecell{77.01\\$\pm$0.26} & \makecell{76.23\\$\pm$0.23} & \makecell{75.21\\$\pm$0.22} & \makecell{71.93\\$\pm$0.24} & \makecell{76.29\\$\pm$0.25} & \makecell{73.50\\$\pm$0.23} & \makecell{78.23\\$\pm$0.23} & \makecell{74.60\\$\pm$0.21}\\
 \hline
 GELU & \makecell{57.38\\$\pm$0.25} & \makecell{47.26\\$\pm$0.29}  & \makecell{71.90\\$\pm$0.24} & \makecell{77.47\\$\pm$0.27} &   \makecell{75.60\\$\pm$0.23} &  \makecell{74.18\\$\pm$0.25} & \makecell{76.88\\$\pm$0.24} & \makecell{75.63\\$\pm$0.25} & \makecell{75.30\\$\pm$0.20} & \makecell{70.56\\$\pm$0.26} & \makecell{75.27\\$\pm$0.23} & \makecell{72.15\\$\pm$0.23} & \makecell{77.21\\$\pm$0.21} & \makecell{73.99\\$\pm$0.24}\\
 \hline
 PAU & \makecell{57.69\\$\pm$0.24} & \makecell{47.40\\$\pm$0.27}  & \makecell{71.56\\$\pm$0.26} & \makecell{77.32\\$\pm$0.27} &   \makecell{75.92\\$\pm$0.22} &  \makecell{74.36\\$\pm$0.24} & \makecell{76.69\\$\pm$0.26} & \makecell{75.86\\$\pm$0.25} & \makecell{75.10\\$\pm$0.20} & \makecell{71.10\\$\pm$0.24} & \makecell{75.79\\$\pm$0.25} & \makecell{72.70\\$\pm$0.26} & \makecell{77.50\\$\pm$0.25} & \makecell{74.10\\$\pm$0.20}\\

 \hline
 Sqish & \makecell{\textbf{61.09}\\$\pm$0.24} & \makecell{47.00\\$\pm$0.24} & \makecell{\textbf{73.10}\\$\pm$0.20} & \makecell{\textbf{78.66}\\$\pm$0.22} &   \makecell{\textbf{77.29}\\$\pm$0.18} &  \makecell{\textbf{76.44}\\$\pm$0.20} & \makecell{\textbf{78.12}\\$\pm$0.21} & \makecell{\textbf{77.02}\\$\pm$0.21} & \makecell{\textbf{76.36}\\$\pm$0.20} & \makecell{\textbf{73.36}\\$\pm$0.18} & \makecell{\textbf{77.10}\\$\pm$0.22} & \makecell{\textbf{74.30}\\$\pm$0.21} & \makecell{\textbf{78.50}\\$\pm$0.22} & \makecell{\textbf{75.02}\\$\pm$0.21}\\ 
 \hline

 \end{tabular}
\caption{Comparison between different baseline activations and Sqish on CIFAR100 dataset. Top-1 accuracy(in $\%$) for a mean of 10 different runs has been reported. mean$\pm$std is reported in the Table.} 
\label{tab3}
\end{center}
\end{table*}

\begin{table*}[!htbp]
\begin{center}
\scriptsize
\begin{tabular}{ |c|c|c|c|c|c|c|c|c|c|c|c|c|c|c|c|c| }
 \hline
\makecell{Activation\\ Function} & AN & LN & \makecell{VGG\\16} & \makecell{WRN\\ 28-10} & \makecell{RN\\50} & \makecell{PARN\\34} & \makecell{DN\\121}  & \makecell{IN\\V3} & \makecell{MN\\V2} & \makecell{SF-V2\\ 2.0x} & RNxt & XT & \makecell{EN\\B0} & \makecell{RN\\18}  \\

 \hline 
 ReLU & \makecell{84.15\\$\pm$0.18} & \makecell{75.87\\$\pm$0.18}& \makecell{93.50\\$\pm$0.20} & \makecell{95.25\\$\pm$0.18} &   \makecell{94.42\\$\pm$0.19} &  \makecell{94.10\\$\pm$0.18} & \makecell{94.68\\$\pm$0.17} & \makecell{94.22\\$\pm$0.22} & \makecell{94.22\\$\pm$0.19} & \makecell{91.65\\$\pm$0.24} & \makecell{93.35\\$\pm$0.19} & \makecell{90.56\\$\pm$0.24} & \makecell{95.12\\$\pm$0.17} & \makecell{94.17\\$\pm$0.22}\\ 
 \hline
 Leaky ReLU & \makecell{84.20\\$\pm$0.20} & \makecell{75.99\\$\pm$0.18} & \makecell{93.60\\$\pm$0.18} & \makecell{95.10\\$\pm$0.21} &   \makecell{94.40\\$\pm$0.18} &  \makecell{94.25\\$\pm$0.20} & \makecell{94.80\\$\pm$0.20} & \makecell{94.11\\$\pm$0.19} & \makecell{94.25\\$\pm$0.16} & \makecell{91.70\\$\pm$0.20} & \makecell{93.30\\$\pm$0.19} & \makecell{90.80\\$\pm$0.24} & \makecell{95.30\\$\pm$0.17} & \makecell{94.10\\$\pm$0.24}\\
 \hline
 PReLU & \makecell{84.36\\$\pm$0.21} & \makecell{75.89\\$\pm$0.22} & \makecell{93.18\\$\pm$0.20} & \makecell{94.94\\$\pm$0.22} &   \makecell{94.21\\$\pm$0.24} &  \makecell{94.28\\$\pm$0.26} &
 \makecell{94.45\\$\pm$0.22} & 
 \makecell{94.41\\$\pm$0.24} & 
 \makecell{94.38\\$\pm$0.20} & \makecell{91.75\\$\pm$0.22} & \makecell{93.39\\$\pm$0.20} & \makecell{91.15\\$\pm$0.23} & \makecell{95.32\\$\pm$0.18} & \makecell{94.20\\$\pm$0.25}\\ 
  \hline
 ELU & \makecell{84.75\\$\pm$0.21} & \makecell{75.82\\$\pm$0.20} & \makecell{93.68\\$\pm$0.120} & \makecell{95.21\\$\pm$0.18} &   \makecell{94.24\\$\pm$0.25} &  \makecell{94.21\\$\pm$0.25} & \makecell{94.51\\$\pm$0.19} & \makecell{94.50\\$\pm$0.20} & \makecell{94.21\\$\pm$0.19} & \makecell{91.88\\$\pm$0.22} & \makecell{93.50\\$\pm$0.16} & \makecell{91.39\\$\pm$0.20} & \makecell{95.43\\$\pm$0.15} & \makecell{94.20\\$\pm$0.22}\\
 \hline
 
 Swish & \makecell{85.34\\$\pm$0.18} & \makecell{\textbf{77.93}\\$\pm$0.18} & \makecell{93.69\\$\pm$0.19} & \makecell{95.34\\$\pm$0.18} &  \makecell{94.59\\$\pm$0.22} &  \makecell{94.57\\$\pm$0.24} & \makecell{94.69\\$\pm$0.20} & \makecell{94.43\\$\pm$0.18} & \makecell{94.48\\$\pm$0.17} & \makecell{92.24\\$\pm$0.22} & \makecell{93.69\\$\pm$0.18} & \makecell{91.78\\$\pm$0.20} & \makecell{95.63\\$\pm$0.16} & \makecell{94.035\\$\pm$0.22}\\
 
 \hline
 Mish & \makecell{85.85\\$\pm$0.18} & \makecell{77.85\\$\pm$0.17} & \makecell{93.90\\$\pm$0.16} & \makecell{95.25\\$\pm$0.16} &  \makecell{94.70\\$\pm$0.20} &  \makecell{94.60\\$\pm$0.20} & \makecell{95.14\\$\pm$0.17} & \makecell{94.70\\$\pm$0.17} & \makecell{94.70\\$\pm$0.20} & \makecell{92.50\\$\pm$0.18} & \makecell{93.90\\$\pm$0.17} & \makecell{92.11\\$\pm$0.22} & \makecell{95.61\\$\pm$0.15} & \makecell{94.60\\$\pm$0.23}\\
 \hline
 GELU & \makecell{85.12\\$\pm$0.20} & \makecell{77.39\\$\pm$0.18} &  \makecell{93.70\\$\pm$0.22} &  \makecell{95.33\\$\pm$0.24} & \makecell{94.88\\$\pm$0.20} & \makecell{94.67\\$\pm$0.18} & \makecell{94.47\\$\pm$0.19} & \makecell{94.35\\$\pm$0.18} & \makecell{94.30\\$\pm$0.17} & \makecell{92.39\\$\pm$0.18} & \makecell{93.55\\$\pm$0.19} & \makecell{91.92\\$\pm$0.21} & \makecell{95.42\\$\pm$0.16} & \makecell{94.39\\$\pm$0.21}\\
 \hline
PAU & \makecell{85.25\\$\pm$0.21} & \makecell{77.69\\$\pm$0.21} & \makecell{93.50\\$\pm$0.21} & \makecell{95.19\\$\pm$0.18} &  \makecell{94.60\\$\pm$0.20} &  \makecell{94.39\\$\pm$0.22} & \makecell{94.77\\$\pm$0.18} & \makecell{94.50\\$\pm$0.17} & \makecell{94.41\\$\pm$0.16} & \makecell{92.22\\$\pm$0.18} & \makecell{93.59\\$\pm$0.17} & \makecell{92.09\\$\pm$0.21} & \makecell{95.41\\$\pm$0.17} & \makecell{94.31\\$\pm$0.21}\\
 \hline
Sqish & \makecell{\textbf{86.79}\\$\pm$0.18} & \makecell{77.36\\$\pm$0.20} & \makecell{\textbf{94.51}\\$\pm$0.11} & \makecell{\textbf{95.71}\\$\pm$0.12} &  \makecell{\textbf{95.32}\\$\pm$0.15} &  \makecell{\textbf{95.07}\\$\pm$0.18} & \makecell{\textbf{95.68}\\$\pm$0.17}  & \makecell{\textbf{95.32}\\$\pm$0.14} & \makecell{\textbf{95.40}\\$\pm$0.12} & \makecell{\textbf{93.79}\\$\pm$0.17} & \makecell{\textbf{94.36}\\$\pm$0.19} & \makecell{\textbf{92.79}\\$\pm$0.21} & \makecell{\textbf{96.04}\\$\pm$0.14} & \makecell{\textbf{94.96}\\$\pm$0.18}\\ 
 
 \hline
  
 \end{tabular}
\caption{Comparison between different baseline activations and Sqish on CIFAR10 dataset. Top-1 accuracy(in $\%$) for a mean of 10 different runs has been reported. mean$\pm$std is reported in the Table.} 
\label{tab2}
\end{center}
\end{table*}
\subsubsection{Tiny ImageNet}

In this section, we delve into the intricacies of image classification using the challenging Tiny Imagenet dataset \cite{tiny}. Tiny Imagenet presents a rich tapestry of $64 \times 64$ RGB visuals, encompassing 1,00,000 training snapshots, 10,000 for validation, and another 10,000 designated for testing, all distributed across 200 diverse classes. To enhance the robustness of our model, we incorporated augmentation strategies, including rotation and horizontal flipping. Our experimental setup involves a batch size set to 64 and an initial learning rate of 0.1. Notably, we apply a meticulous reduction in this rate by a factor of 10 after every successive 50 epochs. The optimization landscape is carved using stochastic gradient descent (SGD) \cite{sgd1, sgd2}, complemented by a 0.9 momentum and a weight decay of $5e^{-4}$. Each network was rigorously trained over 200 epochs. For a comprehensive view, Table~\ref{tabtiny} presents the top-1 accuracy, representing an average over five distinct runs.

\begin{table*}[!t]
\centering
\begin{tabular}{ |c|c|c|c|c|c|c|c|c|}
 \hline
 Activation Function & ResNet-18 & ResNet-50 & \makecell{WideResNet \\ 28-10}\\
 \hline
 ReLU  &  59.21 $\pm$ 0.40 & 61.11 $\pm$ 0.42 &  63.60 $\pm$ 0.37\\ 
 
  \hline
 Leaky ReLU  & 59.31 $\pm$ 0.41 & 61.41 $\pm$ 0.41 &  63.49 $\pm$ 0.39\\ 

 \hline
 PReLU  & 59.69 $\pm$ 0.39 & 61.42 $\pm$ 0.41 &  63.71 $\pm$ 0.42\\ 
 \hline
 ELU   & 59.28 $\pm$ 0.41 & 61.34 $\pm$ 0.40 &  63.79 $\pm$ 0.41\\ 
  \hline
 Swish   & 60.21 $\pm$ 0.41 & 61.69 $\pm$ 0.40 &  64.50 $\pm$ 0.38\\ 
 \hline
 Mish   & 60.32 $\pm$ 0.36 & 62.20 $\pm$ 0.40 &  64.81 $\pm$ 0.36\\ 
 \hline
 GELU   & 60.20 $\pm$ 0.36 & 61.89 $\pm$ 0.40 &  64.32 $\pm$ 0.35\\ 
 \hline
PAU  & 60.51 $\pm$ 0.37 & 61.81 $\pm$ 0.40 &  64.51 $\pm$ 0.36\\ 
 \hline
Sqish & \textbf{61.20} $\pm$ 0.36 & \textbf{63.05} $\pm$ 0.38 &  \textbf{65.92} $\pm$ 0.33\\ 
  \hline
 \end{tabular}
 
\caption{Comparison between Squish and other baseline activations on Tiny ImageNet dataset for image classification problem. We report Top-1 test accuracy (in \%) for the mean of 5 different runs. mean$\pm$std is reported in the Table.} 
\label{tabtiny}
\end{table*}

\subsection{Object Detection}
In this section, we provide experimental results for the object detection task on the Pascal VOC dataset \cite{pascal} using the Single Shot MultiBox Detector (SSD) 300 model \cite{ssd} and consider VGG16 (with batch-normalization) \cite{vgg} as the backbone network. VOC2007 and VOC2012 are used as train data, and VOC2007 is used as the test dataset. There are 20 distinct objects in the dataset. We assume a batch size of 8 and an initial learning rate of 0.001. We employ the SGD \cite{sgd1, sgd2} optimizer with 0.9 momentum and $5e^{-4}$ weight decay, as well as trained networks with up to 120000 iterations. We do not take into account any pre-trained weight. Table~\ref{tabod} shows the mean average accuracy (mAP) for the mean of five separate runs. 
\begin{table}[!htbp]
\centering
\begin{tabular}{ |c|c|c| }
 \hline
 Activation Function &  \makecell{mAP}  \\
 
 \hline
 ReLU  &  77.3 $\pm$ 0.13  \\ 
 \hline
 
 Leaky ReLU &  77.2 $\pm$ 0.15 \\
 \hline
 PReLU & 77.3 $\pm$ 0.14\\
 \hline
 ELU & 77.3 $\pm$ 0.15\\
 \hline
Swish  &  77.5 $\pm$ 0.12\\
 \hline
Mish & 77.6 $\pm$ 0.12\\
 
 \hline
GELU & 77.5 $\pm$ 0.14\\
 \hline
PAU  & 77.5 $\pm$ 0.13\\
\hline
Sqish &  \textbf{78.1} $\pm$ 0.09 \\ 
\hline
 \end{tabular}
\caption{Comparison between Sqish and other baseline activations on Pascal VOC dataset for object detection problem. We report mAP for the mean of 5 different runs. mean$\pm$std is reported in the Table.} 
\label{tabod}
\end{table}

\subsection{3D Medical Imaging}
In the following subsections, we report detailed experimental results on 3D medical image classification and 3D medical image segmentation problems.
\subsubsection{3D Medical Image Classification}
This section presents experimental results for the 3D image classification problem on MosMed dataset \cite{mosmed}. The dataset contains CT scans with COVID-19-related findings (CT1-CT4) and without any findings (CT0). The dataset has 1110 studies. We consider 70\% of the data as training data and 30\% of the data used for testing. We consider a batch size of 8, 0.0001 initial learning rate, and decay the learning rate with cosine annealing learning rate scheduler \cite{cosan}. We use Adam optimizer \cite{adam}, $5e^{-4}$ weight decay, and trained up to 200 epochs with 3D ResNet-18 model. Experimental results are reported in Table~\ref{tabmed}.

\begin{table}[!htbp]
\centering
\begin{tabular}{ |c|c|c| }
 \hline
 Activation Function &  \makecell{Accuracy}  \\
 
 \hline
 ReLU  &  79.50  \\ 
 \hline
 
 Leaky ReLU &  79.67  \\
 \hline
 PReLU & 79.76 \\
 \hline
 ELU & 79.89 \\
 \hline
Swish  & 79.99 \\
 \hline
Mish & 80.17\\
 
 \hline
GELU & 79.96\\
 \hline
PAU  & 80.27\\
\hline
Sqish & \textbf{80.58} \\ 
\hline
 \end{tabular}
\caption{Comparison between Sqish and other baseline activations on MosMedData dataset for 3D image classification with ResNet-18 model. We report Top-1 accuracy for the mean of 5 different runs.} 
\label{tabmed}
\end{table}
\subsubsection{3D Medical Image Segmentation}
This section presents the experimental outcomes for 3D brain tumor segmentation using 3D-UNet \cite{3dunet} on the BraTS 2020 dataset \cite{menze2014multimodal, bakas2017advancing, bakas2018identifying}. This data set contains 369 samples for training and 125 samples for validation. In this experiment, we consider a batch size of 2, and the learning rate is 0.001, Adam optimizer \cite{adam} with $5e^{-4}$ weight decay and cosine annealing learning rate scheduler for training the network. Moreover, we trained this 3D model for 150 epochs. We presented the network performance analysis for Sqish and the baseline activation functions with this network in Table~\ref{tabseg} in terms of Accuracy and Dice Score to measure the performance. 
\begin{table}[!htbp]
\begin{center}
\begin{tabular}{ |c|c|c|c| }
 \hline
 Activation Function & Accuracy &  Dice Score \\
  
 \hline
 ReLU  & 95.04 & 97.37\\ 
  \hline
  Leaky ReLU & 94.98 & 97.34\\
 \hline
  
 PReLU &  94.95 & 97.40\\
 \hline
 ELU &  95.10  &  97.42\\
 \hline
  Swish  &  95.17 & 97.50\\
  \hline
 Mish &  95.12 &  97.48\\
\hline
 GELU &  95.21 &  97.30\\
 \hline
 PAU & 95.27 & 97.54\\
 \hline
 Sqish & 95.45 &  97.68 \\ 
\hline
\end{tabular}
\caption{Comparison between Sqish and other baseline activations on BraTS 2020 Dataset for 3D image segmentation with 3D-Unet Model.} 
\label{tabseg}
\end{center}
\end{table}

\subsection{Adversarial Attack}

In our investigation of adversarial attack challenges, we utilized the CIFAR10 and CIFAR100 datasets, coupling them with the FGSM attack methodology \cite{fgsm}. These datasets were processed with a batch size of 128 and an initial learning rate of 0.01. As we navigated through our experiment, the learning rate underwent a nuanced decay, facilitated by the cosine annealing learning rate scheduler \cite{cosan}. The optimization frontier was led by the stochastic gradient descent \cite{sgd1, sgd2}, fortified with a momentum of 0.9 and a weight decay setting of $5e^{-4}$. Throughout this scientific endeavor, every network was diligently trained over a span of 200 epochs.

For the discerning reader, we've tabulated the Top-1 accuracy in Table~\ref{tabat} (for CIFAR10) and Table~\ref{tabatc10} (for CIFAR100). These results represent an average derived from 10 distinct runs. Our research spotlight was on the robust frameworks of ShuffleNet V2 (2.0x) \cite{shufflenet} and MobileNet V2 \cite{mobile}. 
\begin{table}[!htbp]
\scriptsize
\centering
\begin{tabular}{ |c|c|c| }
 \hline
 Activation Function &  ShuffleNet V2 (2.0x) & MobileNet V2  \\
 
 \hline
 ReLU  &  59.26 $\pm$ 0.11 & 68.89 $\pm$ 0.14 \\ 
 \hline
 
 Leaky ReLU &  59.87 $\pm$ 0.12 & 68.81 $\pm$ 0.15\\
 \hline
 PReLU & 60.15 $\pm$ 0.10 &  69.06 $\pm$ 0.13\\
 \hline
 ELU & 60.36 $\pm$ 0.11 &  69.22 $\pm$ 0.14\\
 \hline
Swish  &  65.66 $\pm$ 0.10 & 69.81 $\pm$ 0.12\\
 \hline
Mish & 65.42 $\pm$ 0.12 & 70.10 $\pm$ 0.12\\
 
 \hline
GELU & 64.84 $\pm$ 0.11 & 69.55 $\pm$ 0.14\\
 \hline
PAU  & 66.10 $\pm$ 0.10 & 69.99 $\pm$ 0.14\\
\hline
Sqish &  \textbf{68.47} $\pm$ 0.09 & \textbf{70.68} $\pm$ 0.12\\ 
\hline
 \end{tabular}
\caption{Comparison between Sqish and other baseline activations on CIFAR100 dataset for FGSM Adversarial attack ($\epsilon = 0.04)$. We report Top-1 accuracy for the mean of 5 different runs. mean$\pm$std is reported in the Table.} 
\label{tabat}
\end{table}
\begin{table}[!htbp]
\scriptsize
\centering
\begin{tabular}{ |c|c|c| }
 \hline
 Activation Function &  ShuffleNet V2 (2.0x) & MobileNet V2  \\
 
 \hline
 ReLU  &  87.41 $\pm$ 0.14 &  90.52 $\pm$ 0.15\\ 
 \hline
 
 Leaky ReLU &  87.59 $\pm$ 0.13 & 90.67 $\pm$ 0.12\\
 \hline
 PReLU & 87.85 $\pm$ 0.14 & 90.60 $\pm$ 0.14\\
 \hline
 ELU & 87.92 $\pm$ 0.16 & 90.80 $\pm$ 0.14\\
 \hline
Swish  &  89.69 $\pm$ 0.12 & 91.35 $\pm$ 0.12\\
 \hline
Mish & 89.35 $\pm$ 0.11 & 91.68 $\pm$ 0.13\\
 
 \hline
GELU & 89.02 $\pm$ 0.12 & 91.33 $\pm$ 0.14\\
 \hline
PAU  & 89.94 $\pm$ 0.12 & 91.74 $\pm$ 0.13\\
\hline
Sqish &  \textbf{91.15} $\pm$ 0.12 & \textbf{92.20} $\pm$ 0.12\\ 
\hline
 \end{tabular}
\caption{Comparison between Sqish and other baseline activations on CIFAR10 dataset for FGSM Adversarial attack ($\epsilon = 0.04)$. We report Top-1 accuracy for the mean of 5 different runs. mean$\pm$std is reported in the Table.} 
\label{tabatc10}
\end{table}
 
\begin{table*}[!t]
\newenvironment{amazingtabular}{\begin{tabular}{*{50}{l}}}{\end{tabular}}
\centering
\begin{amazingtabular}
\midrule
Baselines & ReLU & \makecell{Leaky\\ ReLU} & ELU  & PReLU & Swish & Mish & GELU & PAU\\
\midrule
Sqish $>$ \text{Baseline} & \hspace{0.3cm}43 & \hspace{0.3cm}43 & \hspace{0.3cm}43 & \hspace{0.3cm}43 &  \hspace{0.3cm}43 & \hspace{0.3cm}43 & \hspace{0.3cm}43 & \hspace{0.3cm}43 \\
Sqish $=$ \text{Baseline} & \hspace{0.3cm}0 & \hspace{0.3cm}0 & \hspace{0.3cm}0 & \hspace{0.3cm}0  & \hspace{0.3cm}0  & \hspace{0.3cm}0  & \hspace{0.3cm}0 & \hspace{0.3cm}0\\
Sqish $<$ \text{Baseline}  & \hspace{0.3cm}0 & \hspace{0.3cm}0  & \hspace{0.3cm}0 & \hspace{0.3cm}0 & \hspace{0.3cm}2 & \hspace{0.3cm}2 & \hspace{0.3cm}2 & \hspace{0.3cm}2 \\

\midrule
\end{amazingtabular}

  \caption{Baseline table for Sqish. These numbers represent the total number of models in which Sqish underperforms, equals, or outperforms compared with the baseline activation functions}
  \label{tabbase}
\end{table*}
\subsection{Data Augmentation}
In this section, we report results with the MixUp \cite{mixup} augmentation method. Mixup augmentation is a data augmentation method that uses the training data to produce a weighted combination of random image pairings. Mixup augmentation can assist in preventing overfitting, increasing generalization, and strengthening the model against adversarial attacks. We report results with Mixup augmentation on the CIFAR100 dataset with the ShuffleNet V2 \cite{shufflenet} and ResNet-18 \cite{resnet} model on Table~\ref{tabaug}. To train the network, we consider a batch size of 128, 0.01 initial learning rate, and decay the learning rate with cosine annealing learning rate scheduler \cite{cosan}, stochastic gradient descent \cite{sgd1, sgd2} optimizer with 0.9 momentum \& $5e^{-4}$ weight decay, and trained up to 200 epochs. 

\begin{table}[!htbp]
\footnotesize
\begin{center}
\begin{tabular}{ |c|c|c|c| }
 \hline
 Activation Function & ShuffleNet V2 (2.0x) &  ResNet 18 \\
  
 \hline
 ReLU  & 69.18 $\pm$ 0.20 & 73.79 $\pm$ 0.21\\ 
  \hline
  Leaky ReLU & 69.15 $\pm$ 0.20 & 73.90 $\pm$ 0.23\\
 \hline
  
 PReLU & 69.18 $\pm$ 0.22  & 74.10 $\pm$ 0.22\\
 \hline
  ELU & 69.30 $\pm$ 0.20  & 74.21 $\pm$ 0.21\\
 \hline
  Swish  & 72.69 $\pm$ 0.20  & 74.42 $\pm$ 0.21\\
  \hline
 Mish & 72.98 $\pm$ 0.20  & 74.60 $\pm$ 0.21 \\
\hline
 GELU & 72.90 $\pm$ 0.21  & 74.45 $\pm$ 0.22 \\
 \hline
 PAU & 73.39 $\pm$ 0.20 & 74.80 $\pm$ 0.21\\
 \hline
 Sqish & \textbf{74.40} $\pm$ 0.18  &  \textbf{75.85} $\pm$ 0.18 \\ 
\hline
\end{tabular}
\caption{Top-1 Test Accuracy Reported with Mixup Augmentation Method on CIFAR100 Dataset for the Mean of 5 Different Runs. We Report mean$\pm$std in the Table.} 
\label{tabaug}
\end{center}
\end{table}


\section{Baseline Table}
\label{sec:baseline}

We present a detailed experimental summary in Table~\ref{tabbase}. This Table shows the total number of experiments conducted and the total number of cases where the proposed function outperforms, equals, or underperforms compared with the baseline activation functions. From the baseline table, it is clear that Sqish outperforms baseline activations in most cases. 
\section{Computational Time Comparison}
\label{sec:time}
We present the computational Time Comparison for Sqish and the baseline activation functions in this section. The tests are carried out on an NVIDIA Tesla A100 GPU. The results are reported in Table~\ref{tabtime} for Sqish and other baseline activation functions for a 224 $\times$ 224 RGB image in the ResNet-18 model. From the experiment section and Table~\ref{tabtime}, it is clear that there is a trade-off between computational time and model performance when compared to ReLU or its variants, as Sqish is highly non-linear. Also, note that Sqish improves model performance significantly while training time is comparable with smooth activations like Swish, Mish, GELU, \& PAU.
\begin{table}[!htbp]
\centering
\begin{tabular}{ |c|c|c|c| }
 \hline
 \makecell{Activation\\ Function} &  Forward Pass & Backward Pass \\
 
 \hline
 ReLU  &  
3.83 $\pm$ 0.64 $ \mu$s &
5.17 $\pm$ 0.58 $ \mu$s\\ 
 \hline
Leaky ReLU &  
4.32 $\pm$ 0.11 $\mu$s &
5.20 $\pm$ 0.13 $\mu$s
 \\
  \hline
 
 PReLU & 
5.11 $\pm$ 0.32$ \mu$s &
6.15 $\pm$ 0.33 $ \mu$s
\\

  \hline
 ELU  &  
4.41 $\pm$ 0.58 $ \mu$s &
5.40 $\pm$ 0.38 $ \mu$s
\\
\hline
 
 Mish & 
6.71 $\pm$ 1.69 $ \mu$s &
6.58 $\pm$ 0.44 $ \mu$s
\\

 \hline
 GELU & 
5.50 $\pm$ 0.21 $ \mu$s &
7.52 $\pm$ 0.31 $ \mu$s
\\
 \hline
  
 Swish  &  
5.47 $\pm$ 0.14 $ \mu$s &
6.78 $\pm$ 0.26 $ \mu$s
\\

 \hline
  PAU &  
8.93 $\pm$ 1.22 $ \mu$s &
22.03 $\pm$ 2.08 $ \mu$s\\
\hline
 Sqish &  
5.52 $\pm$ 0.15 $ \mu$s 
& 7.89 $\pm$ 0.56 $ \mu$s
 \\ 
\hline
 \end{tabular}

\caption{Runtime comparison for the forward and backward passes for Sqish and other baseline activation functions for a 224$\times$ 224 RGB image in ResNet-18 model.} 
\label{tabtime}
\end{table}


\section{Conclusion}
\label{sec:conclusion}
In this work, we present \textit{Sqish}, a novel smooth activation function based on an approximation of the maximum function. Our proposed function is smooth and can approximate ReLU or its variants very well. We use Sqish as a trainable activation function for our experiments. We show that in a wide variety of datasets on various deep learning tasks, the proposed activation function outperforms existing and conventional activation functions such as ReLU, Swish, GELU, PAU, and others, in the majority of instances, indicating that replacing the hand-crafted function by Sqish activation functions can be useful in deep networks.

\bibliographystyle{unsrtnat}
\bibliography{references}  






\end{document}